\documentclass[runningheads]{llncs}
\usepackage[T1]{fontenc}
\usepackage{graphicx}
\usepackage{cite}
\usepackage{amsmath,amssymb,amsfonts}
\usepackage{algorithmic}
\usepackage{textcomp}
\usepackage[dvipsnames]{xcolor}
\usepackage{tikz}

\usepackage{enumitem}
\usepackage{float}
\usepackage{pgfplots}
\pgfplotsset{compat=1.16}
\usepackage{booktabs}
\renewcommand{\arraystretch}{1.4} 
\usepackage{makecell}
\usepackage{tabularray}
\usepackage{threeparttable}  
\usepackage{multirow}
\usepackage{subcaption}
\usepgfplotslibrary{fillbetween}
\usepackage{amsthm}
\usepackage[colorlinks=true,urlcolor=blue]{hyperref}  
\begin{document}
%
%
%
%

\title{Few Edges Are Enough: Few-Shot Network Attack Detection with Graph Neural Networks \thanks{This is the author's version of the work accepted for publication at IWSEC 2024. The final version is available \href{https://link.springer.com/chapter/10.1007/978-981-97-7737-2_15}{here}.}}
\titlerunning{Few Edges Are Enough}

%
\author{Tristan Bilot\inst{1,2,3} \and
Nour El Madhoun\inst{3,4} \and
Khaldoun Al Agha\inst{1} \and
Anis Zouaoui \inst{2}}
%
%
\institute{
Université Paris-Saclay, CNRS, LISN,
\email{tristan.bilot@universite-paris-saclay.fr} \\
\email{alagha@lisn.fr} \and
Iriguard,
\email{anis.zouaoui@adservio.fr} \and
LISITE Laboratory, Isep,
\email{nour.el-madhoun@isep.fr} \and
Sorbonne Université, CNRS, LIP6,
}
\maketitle              

\theoremstyle{definition} 
\newtheorem*{ssl}{Self-Supervised Learning (SSL)}
\newtheorem*{dgi}{Deep Graph Infomax (DGI)}
\newtheorem*{few}{Few-Shot Learning}
\newtheorem*{cse}{NF-CSE-CIC-IDS2018-v2}
\newtheorem*{uns}{NF-UNSW-NB15-v2}

\begin{abstract}
Detecting cyberattacks using Graph Neural Networks (GNNs) has seen promising results recently. Most of the state-of-the-art models that leverage these techniques require labeled examples, hard to obtain in many real-world scenarios. To address this issue, unsupervised learning and Self-Supervised Learning (SSL) have emerged as interesting approaches to reduce the dependency on labeled data. Nonetheless, these methods tend to yield more anomalous detection algorithms rather than effective attack detection systems.
This paper introduces \textit{Few Edges Are Enough} (\texttt{FEAE}), a GNN-based architecture trained with SSL and Few-Shot Learning (FSL) to better distinguish between false positive anomalies and actual attacks.
To maximize the potential of few-shot examples, our model employs a hybrid self-supervised objective that combines the advantages of contrastive-based and reconstruction-based SSL. By leveraging only a minimal number of labeled attack events, represented as attack edges, \texttt{FEAE} achieves competitive performance on two well-known network datasets compared to both supervised and unsupervised methods. Remarkably, our experimental results unveil that employing only 1 malicious event for each attack type in the dataset is sufficient to achieve substantial improvements. \texttt{FEAE} not only outperforms self-supervised GNN baselines but also surpasses some supervised approaches on one of the datasets.
\end{abstract}

\section{Introduction}

In our interconnected and digitalized world, cybersecurity is of utmost importance, impacting our society and technological infrastructure. As technology advances rapidly, cyber threats become more sophisticated, emphasizing the need for strong defense systems to combat these evolving challenges. 

With the recent advancements in Machine Learning and Deep Learning, researchers are increasingly turning to these techniques to reinforce computer systems against complex attacks. The abundance of data in network environments makes Deep Learning an attractive approach to uncover hidden patterns within malicious communications occurring in the network.
The emergence of Graph Deep Learning and Graph Neural Networks (GNNs) has generated significant interest in fields where interconnected data is prevalent. Cybersecurity researchers have notably adopted these graph-based models to detect attacks in network data \cite{lo2023xg,pujol2022unveiling,bilot2023graph}, which inherently possess interconnected structures. Consequently, multiple variants of message-passing GNNs \cite{gilmer2017neural} have been successfully applied to diverse network datasets, including network flows, packets and authentication logs, to detect specific attacks such as lateral movements, DDoS, and BotNet attacks, relying heavily on labeled training examples.

However, the reliance on numerous labeled attacks poses challenges for real-world scenarios, where the network topology differs from the labeled dataset. Consequently, researchers have been increasingly exploring unsupervised and self-supervised techniques that do not necessitate labeled attacks in advance. While these methods excel at detecting anomalies and unusual events, their application to attack detection may lead to a high number of false positives on highly imbalanced real-world data. Indeed, the model might struggle to differentiate between an actual attack and an unprecedented action, such as a first-time SSH connection between two hosts, when it lacks any attack training label.

Therefore, we encounter a dual trade-off between the complexity of obtaining a multitude of labeled attacks and the requirement of having some labels to effectively distinguish between anomalies and attacks. To address this trade-off, we propose a viable solution using Few-Shot Learning (FSL), employing a minimal number of labeled examples in conjunction with the Self-Supervised Learning (SSL) loop. Motivated by this idea, we introduce \textit{Few Edges Are Enough} (\texttt{FEAE}), a GNN-based detection system that leverages FSL to more accurately identify attacks while relying on only a very small number of labeled examples. \texttt{FEAE} employs contrastive learning along with a few-shot aware reconstruction-based objective, allowing it to cluster similar unlabeled attack edges while requiring only few malicious edges.

This paper introduces the following contributions:
\begin{itemize}
    \item We present, to the best of our knowledge, the first approach leveraging GNNs with few-shot learning for network attack detection. Other similar works either rely on supervised, self-supervised or fully unsupervised methods.
    \item We introduce a few-shot aware reconstruction loss, which successfully integrates the knowledge of few-shot malicious samples within the self-supervised training of the GNN encoder. This capability allows the model to directly cluster various attack families within the embedding space, enabling the downstream decoder to distinctly separate them from benign activities.
    \item Our experimental findings reveal that the performance of fully-supervised methods can be approached or even exceeded by using merely one few-shot malicious edge per attack family. This result is encouraging, significantly reducing the need for extensive labeling.
\end{itemize}

The paper is organized as follows: Section \ref{sec:background} presents some related works. Section \ref{sec:feae} introduces the architecture of \texttt{FEAE}. Section \ref{sec:datasets} presents the model configurations, used datasets and baselines, along with a performance analysis of \texttt{FEAE}. Section \ref{sec:results} benchmarks the performance of the model compared to the baselines and discusses future research directions, whereas Section \ref{sec:conclusion} concludes this work.

\section{Related Works} \label{sec:background}
Since the rise of Graph Neural Networks (GNNs) across various disciplines, they have proven to be highly effective when applied to different network datasets. In most cases, the input graph is directly constructed from the network topology, where nodes represent hosts characterized by their IP addresses, and edges represent network flows, often associated with features obtained through monitoring tools.

E-GraphSAGE \cite{lo2022graphsage} is a model inspired by GraphSAGE \cite{hamilton2017inductive}, specifically designed for edge-level tasks on network graphs. The message-passing function has been adapted to incorporate edge features before updating the node embeddings, with edge embeddings obtained by concatenating connected pairs of nodes. Due to its simplicity and remarkable performance, this model has become a fundamental building block for numerous other works.

In another study \cite{lan2022minbatch}, authors introduced an improvement to E-GraphSAGE by implementing a pre-sampling step before training. This technique reduces graph size, enhancing scalability, and has shown slight performance improvements on the UNSW-NB15 dataset \cite{moustafa2015unsw}.  

Another innovative approach, E-ResGAT \cite{chang2021graph}, operates on a line graph  instead of an edge-attributed graph. Consequently, the detection task transforms into a node classification problem, where edges and their features are converted into nodes. The GAT model \cite{velivckovic2017graph} serves as the GNN encoder to derive node embeddings, and residual connections are incorporated for propagating original edge features. To achieve greater scalability, the authors propose a neighbor sampling strategy, which uniformly samples neighboring nodes during aggregation.

Beyond these advancements, E-GraphSAGE has also been adapted into a self-supervised method known as Anomal-E \cite{caville2022anomal}. Anomal-E utilizes the same graph structure but does not require any labels during the training of the GNN encoder. Deep Graph Infomax (DGI) \cite{velivckovic2018deep} is employed to generate embeddings for positive and negative graph samples, maximizing local-global mutual information by comparing embeddings to a summarized version of the graph. Negative graphs are created using a corruption function, involving shuffling all edges. E-GraphSAGE functions as the encoder and is trained in a self-supervised manner to distinguish between positive and negative edges. Following training, edge embeddings are classified using an Isolation Forest (IF) \cite{liu2008isolation}, using all benign embeddings as the training set. We elaborate further in Section \ref{sec:feae} on the challenges posed by the requirement of all these benign labels in the classifier and how a Few-Shot Learning (FSL) approach can mitigate this problem.

No specific FSL method exists for GNN-based network attack detection to date. However, FewM-HGCL \cite{liu2022fewm} applies FSL to malware variant detection by transforming binaries into heterogeneous graphs and using contrastive learning with system entity co-occurrence. It employs multiple augmentations to create negative and positive graphs and trains three distinct GAT models on these graphs. A discriminator then evaluates the similarity and dissimilarity between the graphs, and a readout function merges all embeddings for malware classification.

\section{\texttt{FEAE}} \label{sec:feae}
\subsection{Intuition}
Our proposed framework leverages a GNN-based SSL method enhanced with FSL for the detection of network attacks, such as Distributed Denial of Service (DDoS), brute force, and botnets. The primary objective of our framework is to achieve high precision in detecting these attacks using minimal labeling. To this end, the input network flow data are first converted into a large graph representation, where nodes correspond to hosts, identified through their IP addresses, and edges represent the network flows. This graph is enriched with additional flow features, such as the count of packets within a Netflow record and the mean packet size, represented as a vector of edge features.

\texttt{FEAE} comprises three key components: the GNN encoder, the SSL module, and the few-shot decoder. 
The GNN encoder is designed to compute node and edge embeddings, effectively capturing the intrinsic relationships between hosts by leveraging the graph topology and the flow features within the network. It operates by training in a self-supervised manner through the SSL module, which ensures that the generated embeddings preserve the original network structure and semantics. 
The SSL module employs hybrid self-supervised strategies, making use of both contrastive-based and reconstruction-based objectives. The contrastive objective is designed to differentiate between original (positive) and augmented (negative) edges, while the reconstruction objective, combined with few-shot samples, aims to produce dissimilar embeddings for benign and malicious edges identified in the few-shot context, also called malicious few-shot edges. The training of the encoder is conducted end-to-end with the SSL module, enabling the generation of embeddings that are aware of the topology and features associated to malicious activity. 
Lastly, the few-shot decoder is trained separately on the learned edge embeddings. It uses a supervised learning approach focused on edge classification and takes advantage of the few-shot labels related to network attacks. 

\subsection{Comparison with Other Works}
Other SSL techniques, such as Anomal-E \cite{caville2022anomal}, necessitate the inclusion of both benign and attack samples for the GNN encoder, and exclusively benign samples for the Isolation Forest decoder. This indirectly leads to a supervised learning paradigm as it requires prior identification of which samples are benign or malicious. In contrast, our few-shot approach requires only a minimal number of malicious samples, thereby obviating the need for identifying benign samples beforehand.

Moreover, fully self-supervised or unsupervised approaches primarily serve as anomaly detection mechanisms rather than cyberattack detection systems \cite{bilot2023graph,king2023euler,paudel2022pikachu,fang2022lmtracker}. The lack of training labels prevents these models from accurately differentiating between legitimate cyberattacks and false positive anomalies. Conversely, models trained solely under supervised conditions may become excessively tailored to the attack patterns observed during training, limiting their generalizability to unseen attacks. Therefore, FSL is proposed as a balanced intermediary, merging the advantages of both paradigms to enhance the precision in detecting diverse network attacks.

\subsection{Notations}

The subsequent sections of this paper will elaborate on each of the components within \texttt{FEAE} using the scientific notations summarized in Table \ref{tab:notations}.

\begin{table}[htbp]\caption{Notations}
\begin{center}
\footnotesize
\setlength{\tabcolsep}{4pt}
\renewcommand{\arraystretch}{1.2}
\begin{tabular}{r@{\hspace{10pt}}l}
\toprule
$E$  & Training edges\\
$X$  & Training edges' features\\
$k$  & Number of selected few-shot edges in each attack family\\
$m$  & Number of attack families\\
$\mathcal{E}$  & Few-shot edges (benign and malicious)\\
$\mathcal{E}_{\text{mal}}$  & Few-shot edges (only malicious)\\
$Y_\mathcal{E}$  & Few shot edges' labels\\
$\textbf{H}$  & Positive edge embeddings matrix\\
$\widetilde{\textbf{H}}$  & Negative edge embeddings matrix\\

\bottomrule
\end{tabular}
\end{center}
\label{tab:notations}
\end{table}

\subsection{GNN Encoder}
In this paper, we propose a simple and lightweight GNN encoder to perform message-passing between nodes in the network graph. Our proposed GNN encoder first computes node embeddings by leveraging the flow numerical flow features attributed to neighoring edges. Precisely, each node aggregates its neighboring edge features in such a way that both the local neighborhood topology and the flow features are captured. Formally, we compute the aggregation of neighboring edges such that:
\begin{align}
    h_{\mathcal{N}(u)} &= \sum_{v\in \mathcal{N}(u)} e_{uv}, \quad u \in N,
\end{align}
where $e_{uv}$ denotes the feature vector of the edge $(u,v)$, $\mathcal{N}(u)$ represents the neighboring nodes of node $u$, and $h_{\mathcal{N}(u)}$ represents the sum aggregation $u$'s neighboring edges.
We employ here a sum aggregation as it offers interesting injective capabilities. As detailed in Lemma 5 of the Graph Isomorphism Network (GIN) paper \cite{xu2018powerful}, the sum aggregation is characterized by its injectivity concerning node features, meaning that a unique combination of features will produce a distinctive sum. Through the use of a similar sum aggregation on the neighboring edge features, the model effectively preserves a greater amount of the structural features within the graph when contrasted with other aggregations like the mean or maximum.

Following the aggregation of node features, these features are then processed by a linear layer with activation function, as described by the following formula:
\begin{align}
    h_u &= \sigma \left( h_{\mathcal{N}(u)} \textbf{W}_{\text{agg}} \right),
\end{align}
where $h_u$ represents the embedding of node $u$, $\sigma$ is the ReLU activation function, and $\textbf{W}_{\text{agg}}$ is a trainable weight matrix.

Following the calculation of node embeddings, our objective is to generate specific edge embeddings from these node representations. These edge embeddings are intended for the subsequent identification of network attacks, characterized here by malicious edges. To achieve this, embeddings from each pair of connected nodes are concatenated. This concatenated output is then multiplied by a distinct trainable matrix, which aims to learn edge embeddings given the concatenated source and destination node embeddings.
\begin{align}
h_{uv} &= \left[ h_u, h_v \right] \textbf{W}_{\text{edge}},
\end{align}
where $[,]$ designates the concatenation operation and $\textbf{W}_{\text{edge}}$ is the edge-level trainable matrix.

Our experiments have shown that a single-layer of this GNN encoder yields superior results in learning representations from positive and negative edges during the self-supervised phase. 

\subsection{SSL Module}
The SSL technique employed to train the edge encoder plays a crucial role in obtaining meaningful embeddings, essential for effective classification by the few-shot decoder. To address this, we propose a hybrid SSL objective that merges contrastive- and reconstruction-based losses. By integrating both methods, the model gains the ability to distinguish between positive edges and negatively augmented edges, while maximizing the reconstruction error for the malicious few-shot edges. This hybrid approach enhances the discriminative power of the embeddings and contributes to the overall detection performance.

\subsubsection{Contrastive-based Loss}
The recent success of contrastive learning in network intrusion detection \cite{caville2022anomal,bilot2023graph} and many other domains \cite{liu2022graph}, motivated us to adopt it in the SSL training.
Deep Graph Infomax (DGI) is the first attempt to apply contrastive learning to graphs, aiming to maximize the mutual information between patch and global representations of the original graph while minimizing it with negative augmentations. Initially designed for node-level tasks, DGI leverages node features when creating node embeddings. However, Anomal-E has demonstrated the successful application of DGI to edge-level tasks, utilizing edge features. In this version, an edge-level encoder such as E-GraphSAGE, computes the edge embeddings for the original graph $G$ and its negatively augmented version $\widetilde{G}$:
\begin{align}
    \textbf{H} &= \text{enc}(G), \\
    \widetilde{\textbf{H}} &= \text{enc}(\widetilde{G}),
\end{align}
where $\textbf{H}$ and $\widetilde{\textbf{H}}$ correspond to the edge embedding matrix for the original graph and its negative augmentation, respectively. Graph $\widetilde{G}$ is defined by $\widetilde{G} = \mathcal{A}\left(G\right)$ with $\mathcal{A}$ an augmentation function that generates a modified version of $G$. 
The edge embeddings will then be compared to a compact version of the original graph to measure the similarity. This compact version, also known as global summary, is represented by a single vector $\Vec{s}$ that preserves global graph information, achieved through a readout function given by:
\begin{align}
    \Vec{s} &= \sigma\left(\mathcal{R}(\textbf{H})\right),
\end{align}
where $\mathcal{R}$ is the mean readout operation and $\sigma$ is the sigmoid function. For each edge in both the original and augmented graphs, the local-global similarity is measured by calculating the dot product between the corresponding edge and the global summary. Additionally, a weight matrix $\textbf{W}$ is utilized during training to either maximize or minimize these similarities. The sigmoid activation function $\sigma$ is then applied to convert the similarity scores into probabilities, indicating whether the input edge is positive or negative:
\begin{align}
    \mathcal{D}\left(\textbf{H}_{uv}, \vec{s}\right) &= \sigma\left( \textbf{H}_{uv} \textbf{W} \Vec{s} \right), \\
    \mathcal{D}(\widetilde{\textbf{H}}_{uv}, \vec{s}) &= \sigma\left( \widetilde{\textbf{H}}_{uv} \textbf{W} \Vec{s} \right),
\end{align}
where $\mathcal{D}$ is called the discriminator function, which returns the probability of an edge being either positive or negative, whereas $\mathbf{H}_{uv}$ and $\widetilde{\textbf{H}}_{uv}$ respectively represent the positive and negative embeddings for the edge $uv$. 

Ultimately, the encoder is trained using the Binary Cross-Entropy (BCE) loss function on the positive and negative edges:
\begin{multline} \label{eq:loss}
\mathcal{L}_{\text{DGI}} = -\frac{1}{|E|+|\widetilde{E}|}\sum_{uv \in E}  \mathbb{E}_G \left[ \text{log} \mathcal{D}\left( \textbf{H}_{uv}, \vec{s} \right) \right] + \sum_{uv \in \widetilde{E}} \mathbb{E}_{\widetilde{G}} \left[ \text{log} \left( 1- \mathcal{D}\left( \widetilde{\textbf{H}}_{uv}, \vec{s} \right) \right) \right], 
\end{multline}
where $|E|$ and $|\widetilde{E}|$ represent the number of edges in the positive graph and negative graph, respectively. The loss is optimized by comparing the probabilities of positive and negative edges to all-ones and all-zeros vectors. Minimizing this loss enables the model to learn to distinguish between original and fake edges, thus producing meaningful edge embeddings that preserve the original graph information. These embeddings can be used as valuable input vectors for the downstream task, which here is a few-shot classifier.

\subsubsection{Reconstruction-based Loss}
\begin{figure*}
    \centering
    \includegraphics[width=0.95\textwidth]{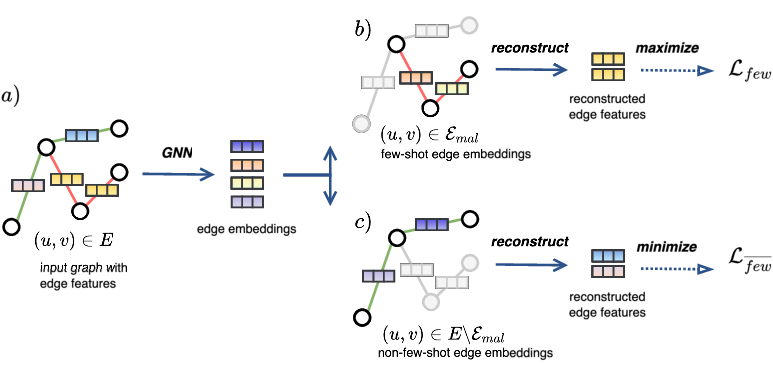}
    \caption{Illustration of the few-shot aware reconstruction-based loss. a) The GNN encoder first compiles features from the local neighborhood edges to produce edge embeddings. Here, red edges represent malicious few-shot edges, whereas green edges symbolize non-few-shot edges, presumed to contain a high rate of benign edges. b) The SSL module leverages the few-shot edges by maximizing the loss associated with these malicious events. This action compels the encoder to create dissimilar edge embeddings for the malicious few-shot edges, ensuring they are easily distinguishable from benign edges. c) The loss function is also designed to minimize the loss for all non-malicious edges.}
    \label{fig:achitecture}
\end{figure*}

While the contrastive-based loss method effectively learns edge representations through self-supervision, it overlooks the potential of leveraging the few-shot labeled samples available in the FSL context. Recognizing the value of these limited labeled edges, we introduce a novel loss function intended to capitalize on these few labeled instances during the SSL phase.
Our goal is to create distinct embeddings specifically for the labeled few-shot examples. To achieve this, we propose an approach inspired by reconstruction-based SSL methods, where we aim to reconstruct edge features from edge embeddings, thereby ensuring that the embeddings of few-shot labeled edges can be easily identified in embedding space. Formally, given an edge embedding $\textbf{H}_{uv}$, the reconstructed edge feature of $(u,v)$ is defined as:
\begin{align}
    \hat{\mathbf{X}}_{uv} = \sigma\left( \mathbf{H}_{uv}\textbf{W}_{\text{rec}} \right)
\end{align}
where $\sigma$ is the sigmoid function and $\textbf{W}_{\text{rec}}$ is a weight matrix with same output dimension as the number of original edge features.
The Mean Squared Error (MSE) loss is used to measure the reconstruction error between the reconstructed and the original edge features. We specifically divide this problem in two separate loss functions $\mathcal{L}_{\text{few}}$ and $\mathcal{L}_{\overline{\text{few}}}$, which correspond to the reconstruction loss of the few-shot and non-few-shot examples, respectively:
\begin{align}
    \mathcal{L}_{\text{few}} &= \sum_{uv \in \mathcal{E}_{\text{mal}}} \left(\mathbf{X}_{uv} - \hat{\mathbf{X}}_{uv}\right)^2, \\
    \mathcal{L}_{\overline{\text{few}}} &= \sum_{uv \in E \backslash \mathcal{E}_{\text{mal}} } \left(\mathbf{X}_{uv} - \hat{\mathbf{X}}_{uv}\right)^2,
\end{align}
where $\mathcal{E}_{\text{mal}}$ is a set of $k$ malicious edges selected among each of the $m$ attack families present in the dataset, $E \backslash \mathcal{E}_{\text{mal}}$ is the set of remaining (unlabeled) non-few-shot edges, and $\mathbf{X}_{uv}$ represents the original features of edge $(u,v)$.
As illustrated in Fig. \ref{fig:achitecture}, our aim is to maximize the reconstruction loss of the $k*m$ malicious few-shot edges while simultaneously minimizing the reconstruction loss of the other non-few-shot edges.
This dynamic of maximizing loss for a few malicious instances while minimizing loss for many benign edges fosters the creation of edge embeddings that are distinctively separable in the embedding space. This separation is achievable provided the dataset exhibits significant imbalance, with the majority of edges being benign, thereby statistically ensuring that non-few-shot edges predominantly consist of benign examples.

The resulting loss function presented in Eq. \ref{eq:our_loss}, integrates the contrastive-based and reconstruction-based objectives presented previously:
\begin{align} \label{eq:our_loss}
\mathcal{L}_{\text{\texttt{FEAE}}} = \mathcal{L}_{\text{DGI}} + \alpha \mathcal{L}_{\overline{\text{few}}} - \beta \mathcal{L}_{\text{few}} 
\end{align}
where $\mathcal{L}_{\text{DGI}}$ is the loss from DGI presented in Eq. \ref{eq:loss}. $\alpha$ and $\beta$ are trade-off coefficients to balance the reconstruction error of few-shot and non-few-shot examples. We recommend to set $\alpha < \beta$, particularly when the dataset contains a significant number of malicious edges. Indeed, $\mathcal{L}_{\overline{\text{few}}}$ should be controlled by $\alpha$ to avoid minimizing the reconstruction error of unlabeled malicious edges present in the non-few-shot edges $E \backslash \mathcal{E}_{\text{mal}}$. Conversely, special attention should be directed towards maximizing $\mathcal{L}_{\text{few}}$ over the malicious few-shot edges, as they have a more significant impact than all other unlabeled edges.





\subsection{Few-Shot Decoder}
The decoder is responsible for the classification of the edge embeddings computed by the encoder and the SSL module; therefore, its training is separate from theirs.
It consists of a 2-layer Multi-Layer Perceptron (MLP) trained using BCE in a supervised manner on the few-shot edge embeddings. The decoder outputs a prediction $\hat{\textbf{y}}$ for every edges such that:
\begin{align}\label{eq:decoder}
\hat{\mathbf{y}} = \sigma\left(\text{MLP}\left(\textbf{H}\right)\right)
\end{align}
where $\sigma$ is the sigmoid function.
To mitigate overfitting on the malicious few-shot edges, we introduce a new set $\mathcal{E}$ that supplements $\mathcal{E}_{\text{mal}}$ such that $\mathcal{E}_{\text{mal}} \subset \mathcal{E}$, by randomly selecting benign edges from the dataset while attempting to maintain the same class distribution. However, in the few-shot scenario, information on labels and their distribution is considered unavailable. As a result, we only select a fixed percentage of benign edges, which varies based on the original dataset distribution. For datasets recognized to be highly imbalanced, primarily consisting of benign samples, as exemplified in this paper, we assume the selection of random edges as the benign few-shot edges for the encoder.\\
The decoder is trained over all edges in $\mathcal{E}$ by computing the BCE loss between the prediction $\hat{y}$ and the actual few-shot label such that:
\begin{align}
\mathcal{L}_{\text{DEC}} = \frac{\sum_{uv \in \mathcal{E}} \text{BCE}\left(\hat{\textbf{y}}_{uv}, \textbf{y}_{uv}\right)}{|\mathcal{E}|},
\end{align}
where $|\mathcal{E}|$ represents the number of select few-shot edges and $\textbf{y}_{uv} \in Y_{\mathcal{E}}$ denotes the label of an edge $(u,v)$, which is set to a malicious label if $(u,v) \in \mathcal{E}_{\text{mal}}$ and a benign label otherwise.


\section{Experiments} \label{sec:datasets}






\subsection{Configurations and Datasets}

The \texttt{FEAE} architecture consists of one encoder layer, described in Section \ref{sec:feae}), trained for a maximum of 600 epochs, with early stopping set to 150 epochs. For all experiments, we utilized the sum aggregation method, as it provided better performance than the mean aggregation, which led to smoother embeddings. Both the encoder and SSL module were implemented with a hidden size of 128 neurons. To train the encoder, we used a learning rate of $1\times10^{-3}$ with the Adam optimizer. Additionally, we applied a weight decay of $1\times10^{-2}$ to the optimizer to better control large gradients generated by the sum aggregation. Regarding the SSL module, the trade-off parameters $\alpha$ and $\beta$ were set to 0.2 and 0.8, respectively, and 5\% of edges in $E$ were randomly selected as benign samples to create $\mathcal{E}$, the full set of selected few-shot edges. The few-shot decoder was trained for a maximum of 4000 epochs, and we applied early stopping at 1500 epochs. We used the Adam optimizer for training, setting the learning rate to $1\times10^{-3}$ and the weight decay to $1\times10^{-5}$.

Our experiments were conducted on two network datasets \cite{sarhan2022towards} commonly used in GNN-based detection methods \cite{bilot2023graph}.

\begin{cse}
This dataset is a Netflow version of the original CSE-CIC-IDS2018 dataset \cite{sharafaldin2018toward}, containing approximately 18.9 million network flows. Among these flows, around 12\% correspond to attack samples, which are divided into 6 attack families including BruteForce, Bot, DoS, DDoS, Infiltration, Web attacks.
\end{cse}
\begin{uns}
Also converted to Netflow format, this version of the UNSW-NB15 dataset \cite{moustafa2015unsw} comprises 2.3 million flows, with attack samples accounting for 4\% of the dataset, distributed across 9 attack families including Fuzzers, Analysis, Backdoor, DoS, Exploits, Generic, Reconnaissance, Shellcode, Worms.
\end{uns}

Both datasets have been standardized using the Netflow format to facilitate ease of use and benchmarking across different detection methods. They offer 43 standardized network flow features, which we employed as edge features in our experiments. Similar to the approach used in Anomal-E, we used only 10\% of each dataset for scalability concerns, with 70\% used as the train set and 30\% as the test set. The GNN encoder and SSL module are trained end-to-end on a graph containing both benign and malicious edges. Subsequently, the few-shot decoder is trained separately using these embeddings to classify edges.

The forward and backward steps were performed on an NVIDIA Tesla V100 GPU with 32GB of memory and an Intel Xeon Gold 6148 CPU with 20 cores and 60GB of memory. The \texttt{FEAE} architecture along with the different baseline models were developed in PyTorch using the DGL library.

\subsection{Performance with Respect to $k$}

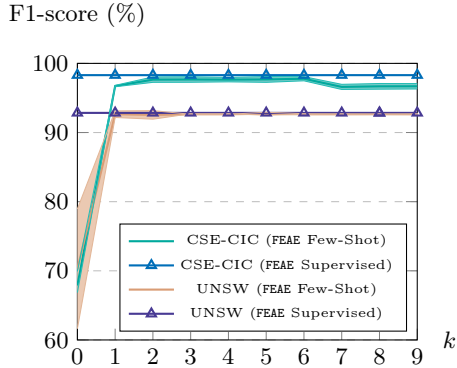
\begin{figure}[!t]
\centering
\begin{tikzpicture}
    \begin{axis}[
      width=0.5\textwidth,
      xmin=0, xmax=9,
      ymin=60, ymax=100,
      ymajorgrids=true,
      grid style=dashed,
      legend pos=south east,
      scatter/use mapped color={draw=black},xlabel={$k$},ylabel={F1-score (\%)},
      xlabel style={at={(1.05,0)},right},
      ylabel style={at={(0,1.1)},above,rotate=-90},
      xtick={0, 1, 2, 3, 4, 5, 6, 7, 8, 9}
    ]

    \addplot [thick, color=Emerald] coordinates {(0,67.88)(1,96.72)(2,97.65)(3,97.67)(4,97.65)(5,97.68)(6,97.78)(7,96.6)(8,96.68)(9,96.69)};
    \addplot [thick, color=NavyBlue, mark=triangle] coordinates {(0,98.3)(1,98.3)(2,98.3)(3,98.3)(4,98.3)(5,98.3)(6,98.3)(7,98.3)(8,98.3)(9,98.3)};

    \addplot [thick, color=Tan] coordinates {(0,70.28)(1,92.65)(2,92.55)(3,92.64)(4,92.64)(5,92.72)(6,92.64)(7,92.64)(8,92.65)(9,92.64)(10,92.64)};
    \addplot [thick, color=BlueViolet, mark=triangle] coordinates {(0,92.84)(1,92.84)(2,92.84)(3,92.84)(4,92.84)(5,92.84)(6,92.84)(7,92.84)(8,92.84)(8,92.84)(9,92.84)};

    \addplot[name path=us_top,color=Emerald!70] coordinates {(0,69.910000000000004)(1,96.82)(2,98.04)(3,98.07000000000001)(4,97.97)(5,98.10000000000001)(6,98.08)(7,96.97)(8,97.09)(9,97.08)};
    \addplot[name path=us_down,color=Emerald!70] coordinates {(0,66.85)(1,96.62)(2,97.26)(3,97.27)(4,97.33000000000001)(5,97.26)(6,97.48)(7,96.22999999999999)(8,96.27000000000001)(9,96.3)};
    \addplot[Emerald!50,fill opacity=0.5] fill between[of=us_top and us_down];

    \addplot[name path=us_top,color=Tan!70] coordinates {(0,61.620000000000005)(1,92.18)(2,91.92999999999999)(3,92.72)(4,92.72)(5,92.51)(6,92.64)(7,92.68)(8,92.67)(9,92.68)(10,92.68)};
    \addplot[name path=us_down,color=Tan!70] coordinates {(0,78.94)(1,93.12)(2,93.17)(3,92.64)(4,92.64)(5,92.92999999999999)(6,92.72)(7,92.68)(8,92.67)(9,92.68)(10,92.68)};
    \addplot[Tan!50,fill opacity=0.5] fill between[of=us_top and us_down];

    \legend{\tiny CSE-CIC (\texttt{FEAE} Few-Shot), \tiny CSE-CIC (\texttt{FEAE} Supervised), \tiny UNSW (\texttt{FEAE} Few-Shot), \tiny UNSW (\texttt{FEAE} Supervised)}
    \end{axis}
\end{tikzpicture}
\caption{\texttt{FEAE} performance with respect to $k$. Setting $k=0$ indicates that only benign edges are used for training, without any labeled malicious edge.}
\label{fig:performance}
\end{figure}

The effectiveness of \texttt{FEAE} across the two presented datasets, with respect to varying $k$ values, is showed in Fig. \ref{fig:performance}. Here, $k$ denotes the count of malicious few-shot edges per each attack family used in training both the SSL module and the few-shot decoder. 
Additionally, we compare the performance to the supervised variants of \texttt{FEAE} that leverage all available supervised examples for training.  The evaluation metric chosen for this analysis is the macro F1-score, presented in Eq. \ref{eq:f1}. This metric is particularly suitable as it provides a balanced measure of the precision and recall across all classes, which is crucial in imbalanced datasets where attack events are significantly less frequent than benign events.
\begin{align}
F1 = 2 \times \frac{{\text{precision} \times \text{recall}}}{{\text{precision} + \text{recall}}}
\label{eq:f1}
\end{align}
This metric takes into account both false positives and false negatives, providing a comprehensive evaluation that is particularly valuable in scenarios where class distributions are skewed.

On the NF-CSE-CIC2018-v2 dataset, the results in the figure indicate a remarkable 96.40\% F1-score with only $k=1$, and the performance further improves to 97.65\% with $k=4$. This demonstrates the ability of the few-shot aware reconstruction loss function to learn effective representations using a minimal number of malicious labels.

For the NF-UNSW-NB15-v2 dataset, a 92.60\% F1-score is reached starting from $k=1$, and the performance remains nearly linear for $k>1$.

In the two outlined scenarios, employing merely one or two malicious samples within this model's framework suffices to reach performance levels comparable to those of fully supervised baselines, which leverage the entirety of available labels. To guarantee a detailed evaluation, the ensuing sections will detail a comparative analysis against various baseline models.

\subsection{Baselines}

To evaluate the performance of \texttt{FEAE}, we conducted a benchmark using multiple GNN baselines that follow different learning paradigms.

\subsubsection{Supervised Baselines}
We introduced four supervised baselines to compare the performance of fully-supervised and few-shot approaches. The first baseline chosen was E-GraphSAGE, as detailed in Section \ref{sec:background}. This model is foundational for malicious edge classification in network graphs. The remaining three baselines—LineGAT, LineGCN, and LineSAGE—are adaptations of the GAT, GCN, and GraphSAGE models, respectively, applied to a line graph context. Given that these GNNs are originally designed for node features, their direct application to scenarios leveraging edge features is impractical. To accommodate this, we converted each edge into a node within a line graph, thus framing it as a node-classification problem where edge features are embodied within nodes. A comparable approach employing a line graph was also utilized in E-ResGAT. However, the complex architecture of this model precluded experiment replication due to memory exhaustion, even when employing the sampling strategy recommended in the original publication. For E-GraphSAGE, a single layer configuration with 128 hidden neurons yielded best results, whereas all line graph-based baselines were implemented with two layers, each comprising 128 hidden neurons.

\subsubsection{Benign-Supervised Baselines}
We also propose multiple variants of the Anomal-E model, as it is the main framework leveraging self-supervised learning to detect network attacks using flow features. Anomal-E consists of the E-GraphSAGE model as the encoder, DGI as the SSL module, and an Isolation Forest (IF) as the decoder. Anomal-E falls under the benign-supervised category as it requires knowledge of all benign edges to train the IF classifier. We evaluate two variants of this model, using different positive and negative augmentations in the SSL training. Findings in \cite{bilot2023benchmark} reveal that employing different augmentation techniques for the positive and negative graphs can improve the prediction accuracy. DGI and Anomal-E both use the original graph as the positive graph and the original graph with randomly permuted edges as the negative graph. However, using different augmentations can improve detection performance of such self-supervised techniques depending on the underlying dataset. The first augmentation, denoted as $\text{aug}_1$, involves using the same random edge permutation as the negative graph, while the positive graph is constructed by randomly dropping 30\% of the edges, along with all the connected edges. This augmentation aims to simulate a graph with fewer hosts while conserving the same topology between the remaining hosts. The second augmentation, denoted as $\text{aug}_2$, employs a positive graph that incorporates new edges with random features, and the negative graph is obtained by randomly masking 30\% of the edges. This second augmentation enforces the model to better generalize to new topologies with different features, while distinguishing between real and noisy features.

\subsubsection{Few-Shot Baselines}
Given the absence of established few-shot GNN baselines for attack detection, we introduce a modified version of Anomal-E. In this adaptation, the IF classifier is substituted with the few-shot classifier from \texttt{FEAE}. The SSL module incorporating DGI remains unaltered to enable a direct comparison with the SSL module employed in our methodology. Moreover, we assess three variations of \texttt{FEAE}: one adheres to the same positive and negative data augmentations as used in DGI, while the other two apply the augmentations discussed earlier. For each baseline, we restrict the use of $k=1$ malicious few-shot edge per attack family, as dictated by the findings presented in Table \ref{fig:performance}. This limitation results in the total utilization of 6 and 9 labeled malicious edges for the NF-CSE-CIC-IDS2018-v2 and NF-UNSW-NB15-v2 datasets, respectively.

\subsection{Experimental Results}

\begin{table*}[ht]
\caption{Performance benchmark between \texttt{FEAE} and other baselines.}
\centering
\scriptsize
\begin{tabular}{@{}llcccccc@{}}\toprule
\textbf{Data} & \textbf{Model} & \multicolumn{3}{c}{\textbf{NF-CSE-CIC-IDS2018-v2}} & \multicolumn{3}{c}{\textbf{NF-UNSW-NB15-v2}} \\
& & \textbf{F1} & \textbf{Precision} & \textbf{Time} & \textbf{F1} & \textbf{Precision} & \textbf{Time} \\
\midrule

$A$, $X$, $Y$ & E-GraphSAGE & $96.02$ & $98.82$ & 0.31 & $95.35$ & $92.49$ & 0.32 \\ 
$A$, $X$, $Y$ & LineGAT & $93.84$ & $96.84$ & 4.3 & $95.33$ & $91.81$ & 14.2 \\
$A$, $X$, $Y$ & LineGCN & $89.29$ & $95.42$ & 0.43 & $95.35$ & $91.83$ & 0.58 \\
$A$, $X$, $Y$ & LineSAGE & $94.94$ & $97.10$ & 1.00 & $\mathbf{95.90}$ & $93.11$ & 2.08 \\
\cmidrule{1-8}
$A$, $X$, $Y_{\text{ben}}$ & Anomal-E (IF) & $94.46$ & $96.86$ & 85.1 & $91.14$ & $85.78$ & 9.2 \\
$A$, $X$, $Y_{\text{ben}}$ & Anomal-E (IF) + $\text{aug}_1$ & $96.53$ & $98.84$ & 81.3 & $87.38$ & $84.13$ & 7.9 \\
\cmidrule{1-8}
$A$, $X$, $Y_{\text{few}}$ & Anomal-E (Few-Shot) & $95.3$ & $97.28$ & 24.5 & $92.47$ & $86.42$ & 1.45 \\
$A$, $X$, $Y_{\text{few}}$ & \texttt{FEAE} & $96.40$ & $99.05$ & 19.6 & $92.64$ & $87.69$ & 1.22 \\
$A$, $X$, $Y_{\text{few}}$ & \texttt{FEAE} + $\text{aug}_1$ & $\mathbf{97.44}$ & $98.71$ & 18.4 & $92.64$ & $87.69$ & 1.19 \\

\bottomrule
\end{tabular}
\begin{tablenotes}
  \setlength\labelsep{0pt}
  \footnotesize
  \item \textbf{Data} refers to the input data required to train a specific \textbf{Model}. The first group of baselines represents supervised approaches, whereas the second group represents benign-supervised methods (with $Y_{\text{ben}}$ corresponding to all benign labels) and the last group is dedicated to the few-shot baselines (with $Y_{\text{few}}$ corresponding to the malicious few-shot labels). \textbf{Time} corresponds to the overall training time in minutes, to reach the score of the corresponding baseline. For both datasets, the performance is measured using the macro \textbf{F1} score and the \textbf{Precision} over 5 iterations, with the notation mean $\pm$ standard deviation.
\end{tablenotes}
\label{table:baselines}
\end{table*}

\begin{figure*}
\centering
\begin{subfigure}{0.5\textwidth}
  \centering
  \includegraphics[width=.95\linewidth]{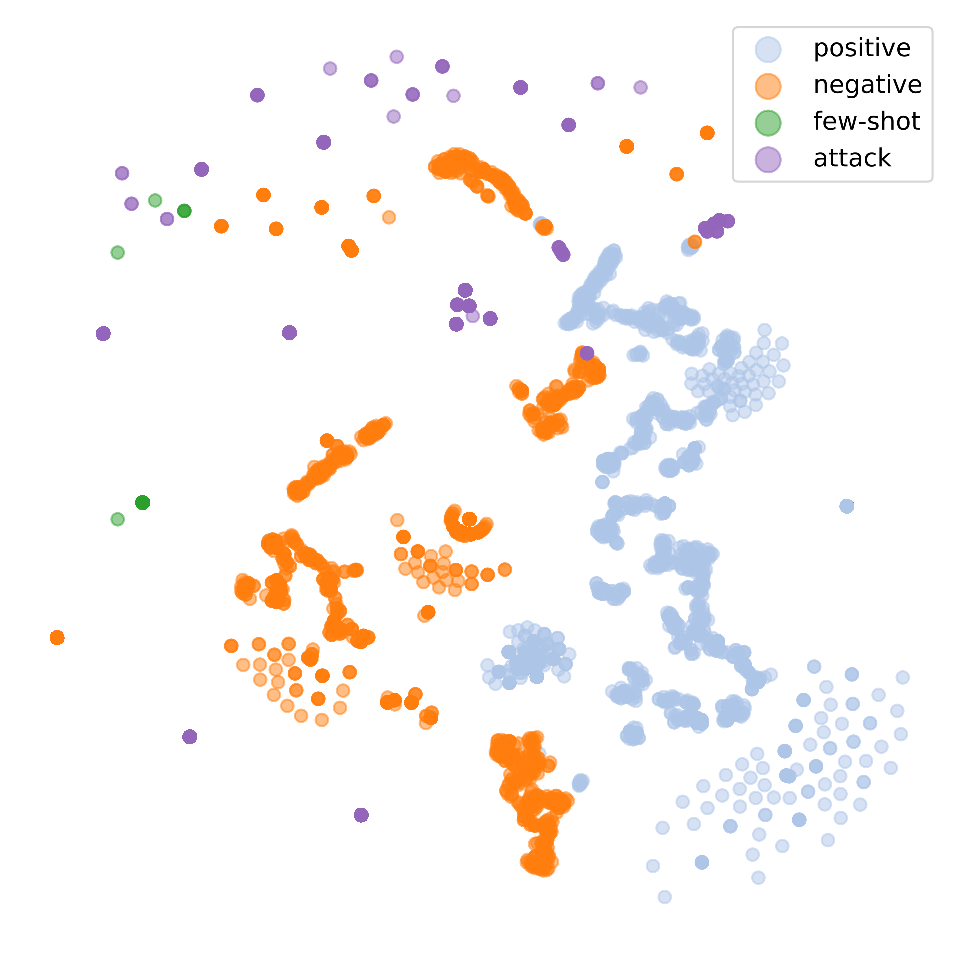}
    \label{fig:enter-label}
\end{subfigure}%
\begin{subfigure}{0.5\textwidth}
  \centering
  \includegraphics[width=.95\linewidth]{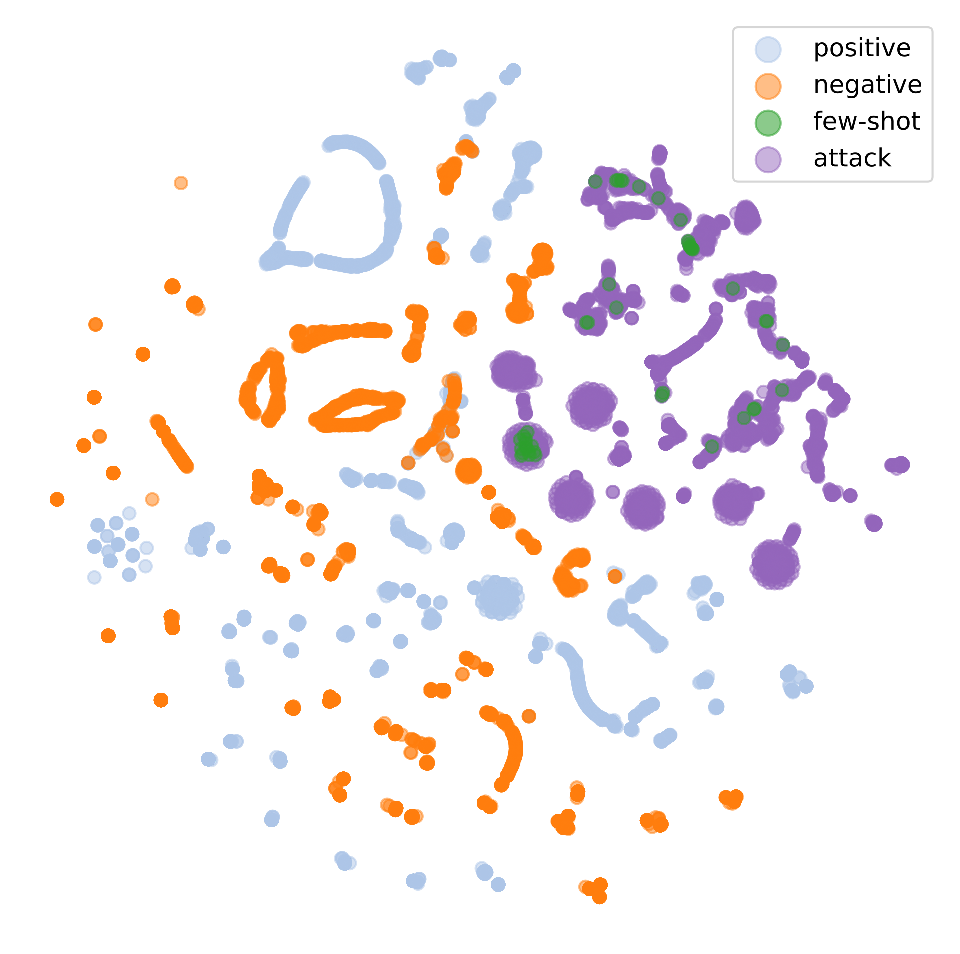}
    \label{fig:enter-label}
\end{subfigure}
\caption{\textbf{Left:} Some edge embeddings produced by Anomal-E. Note that the few-shot edges are just for comparison as they are not leveraged in the original Anomal-E. \textbf{Right:} Edge embeddings generated by \texttt{FEAE}. }
\label{fig:embeddings}
\end{figure*}

The experimental results are summarized in Table \ref{table:baselines}, highlighting the performance of the different approaches described in Section \ref{sec:datasets}.

On the NF-CSE-CIC-IDS2018-v2 dataset, the variant of \texttt{FEAE} employing the first augmentation strategy achieved the highest performance. This result demonstrates the effectiveness of FSL methods within this dataset, as it surpasses the results of fully supervised approaches by utilizing merely one label per attack family. Additionally, \texttt{FEAE} exhibits enhanced performance compared to Anomal-E in the few-shot scenario, highlighting the capabilities of the hybrid SSL objective incorporated in \texttt{FEAE}. 
For a deeper insight into the learning mechanisms of the model, we direct attention to Fig. \ref{fig:embeddings}, which illustrates 5000 edge embeddings produced by the encoder within \texttt{FEAE}.
On the left are embeddings from E-GraphSAGE trained with DGI in Anomal-E, highlighting the model's ability to separate positive and negative edges without specifically focusing on attack edges (purple). On the right, \texttt{FEAE}'s embeddings using DGI's augmentations are displayed, offering insights into its differentiation strategy.
In the case of Anomal-E, the model successfully separates positive and negative edges without giving special attention to attack edges (denoted in purple).
Additionally, attack edges do not cluster under a common region in the embedding space, as the loss function does not specifically require this clustering. However, when looking at the \texttt{FEAE} edge embeddings, we notice a distinct cluster of attack edges, which includes the labeled few-shot edges used during the training of the encoder. By clustering these attack edges directly in the SSL training, \texttt{FEAE} aids the downstream classifier in better learning to differentiate between benign and malicious edges.
Furthermore, in the embeddings from Anomal-E, attack edges do not aggregate into a unified region within the embedding space, due to the absence of a specific requirement in the loss function for such clustering. In contrary, the \texttt{FEAE} edge embeddings exhibit a notable clustering of attack edges. This cluster encompasses the few-shot labeled edges used during the training of the encoder. By directly clustering these attack edges in the SSL training phase, \texttt{FEAE} successfully clusters  benign and malicious edges, which facilitates the classification by the downstream decoder. 

On the NF-UNSW-NB15-v2 dataset, the observed performance exceeds that of Anomal-E yet does not reach the results of supervised baselines. This indicates that the SSL models may not have been optimally trained for this dataset. It highlights a requirement for further experimentation and exploration into the application of SSL methodologies to this particular dataset.

The experiments also highlight the importance of using different graph augmentations, as these techniques lead to an improvement of up to 1\% in the F1-score for \texttt{FEAE} and up to 2\% for Anomal-E. However, we also notice that one augmentation pair may not perform well on both datasets, suggesting that this strategy requires a prior understanding of the underlying data to be effectively applied.

\subsection{Scalability}
The \texttt{FEAE} model demonstrates a significant reduction in computation time compared to the computationally intensive IF classifier used in Anomal-E baselines. Indeed, the IF classifier used in Anomal-E accounts for a significant portion of the computational time, thus serving as a primary bottleneck. Furthermore, experimental evidence suggests that the integration of few-shot examples within the learning process of \texttt{FEAE} reduces the total number of training epochs, as a extended training on the few-shot examples may induce overfitting. This effectively reduces the overall training time compared to fully self-supervised approaches like Anomal-E.

Despite their advantages, both \texttt{FEAE} and Anomal-E incur higher computational costs than simpler supervised GNN models, which benefit from simpler end-to-end training. Indeed, self-supervised models based on DGI are inevitably slower than the supervised ones, which do not require any SSL module. Nonetheless, the trade-off between computational expense and the ability to efficiently handle few-shot scenarios with \texttt{FEAE} justifies its application, particularly in the cybersecurity domain where dataset labeling is scarce.
Efforts to refine \texttt{FEAE}'s architecture could further enhance its computation runtime for applications demanding quick and precise network attack detection with minimal labeled examples.

\section{Results and Discussion} \label{sec:results}

\subsection{Discussion and Future Directions}

The conducted experiments have demonstrated the interesting capability of detecting attacks using only one malicious example per attack class through few-shot learning. In a real-world scenario, this labeling strategy could reduce the reliance on benign-supervised methods, which require the assumption of training the model solely using benign examples. However, it is often challenging to ensure that the labeled examples are genuinely benign, which can be a costly and time-consuming process. Few-shot learning, on the other hand, offers a promising alternative, as it requires only a small number of labeled malicious examples to be effective. 

Furthermore, using graphs in network attack detection offers a significant benefit: cybersecurity analysts can leverage the predictions of the \texttt{FEAE} model to approximate the initial structure of an attack graph. This is achieved by gathering and analyzing the edges that \texttt{FEAE} identifies as malicious. Through this process, analysts can gain insights into the attack's topology and understand how different nodes (potentially representing users, servers, endpoints, or other network entities) are interconnected in a malicious campaign. This capability not only aids in understanding the scope and method of an attack but also in devising targeted countermeasures by revealing the attack's propagation path within the network.

A limitation of \texttt{FEAE} and similar few-shot learning approaches is their dependency on requiring some malicious activity data, which may not exist in the historical data of many enterprise networks.
To address this challenge, future research might focus on integrating malicious activities from synthetic sources or datasets with known malicious content into real-world datasets. Such an approach could reduce the necessity for labeling, as labels from these enriched datasets would suffice. This strategy not only promises to alleviate the labeling burden but also verifies that the performance of the model remains unaffected by new attack types. Moreover, exploring the influence of flow features on prediction accuracy and conducting studies on datasets limited to topological data, devoid of additional features, could provide deeper insights into the model's operational dynamics in varied environments.

Moreover, substantial effort should be dedicated to enhancing the scalability of such models, especially when dealing with large graphs that may undergo structural changes over time. As networks evolve and grow, the model's ability to adapt and handle dynamic structures efficiently becomes crucial. Research into developing scalable approaches for few-shot learning on large and dynamic graphs would greatly benefit practical deployment in real-world environments.

\section{Conclusion} \label{sec:conclusion}
In this study, we undertake the first analysis of applying few-shot learning to network-based attack detection with GNNs.
Through the introduction of the \texttt{FEAE} architecture and the conducted experiments, it has been demonstrated that merely one labeled attack example per attack family is sufficient to achieve competitive detection capabilities when compared to fully-supervised or benign-supervised approaches. These encouraging results open up exciting possibilities for reducing the reliance on fully supervised methods and addressing the challenges of limited labeled data in the field of attack detection. The integration of few-shot learning in attack detection represents a promising research direction, offering more efficient and effective solutions that can adapt well to real-world scenarios.

%
%
\bibliographystyle{splncs04}
\bibliography{biblio}

\end{document}